\documentclass[runningheads]{llncs}
\usepackage[springer]{definition}
\usepackage{graphicx}
\usepackage{subfig}
\usepackage{booktabs,caption}
\usepackage{threeparttable}
\usepackage{algorithm}
\usepackage{algpseudocode}
\usepackage{tabularx}
\usepackage{multirow}
\usepackage{ltablex}
\usepackage{hyperref}

\newcommand{\ie}[0]{{i.e.}}

\newcommand{\imgedit}[0]{\texttt{DNE}}
\newcommand{\gezo}[0]{\texttt{GeZO}{}}

\begin{document}
\title{Debiased Noise Editing on Foundation Models for Fair Medical Image Classification}
%
%
\author{Ruinan Jin\inst{1}\inst{2} \and
Wenlong Deng\inst{1}\inst{2} \and
Minghui Chen\inst{1}\inst{2} \and Xiaoxiao Li\inst{1}\inst{2}}

%
\authorrunning{Jin, R. et al.}
%
\institute{
\textsuperscript{1} The University of British Columbia, Vancouver, BC V6Z 1Z4, Canada\\
\textsuperscript{2} Vector Institute, Toronto, ON M5G 1M1, Canada\\
\email{ruinanjin@alumni.ubc.ca, \{dwenlong,chenmh\}@student.ubc.ca xiaoxiao.li@ece.ubc.ca}
}

%
\maketitle              
\begin{abstract}
In the era of Foundation Models' (FMs) rising prominence in AI, our study addresses the challenge of biases in medical images while the model operates in black-box (e.g., using FM API), particularly spurious correlations between pixels and sensitive attributes. Traditional methods for bias mitigation face limitations due to the restricted access to web-hosted FMs and difficulties in addressing the underlying bias encoded within the FM API. We propose a D(ebiased) N(oise) E(diting) strategy, termed \imgedit, which generates \imgedit{} noise to mask such spurious correlation. \imgedit{} is capable of mitigating bias both within the FM API embedding and the images themselves. Furthermore, \imgedit{} is suitable for both white-box and black-box FM APIs, where we introduced G(reedy) (Z)eroth-O(rder) (\gezo) optimization for it when the gradient is inaccessible in black-box APIs. Our whole pipeline enables fairness-aware image editing that can be applied across various medical contexts without requiring direct model manipulation or significant computational resources. Our empirical results demonstrate the method's effectiveness in maintaining fairness and utility across different patient groups and diseases. In the era of AI-driven medicine, this work contributes to making healthcare diagnostics more equitable, showcasing a practical solution for bias mitigation in pre-trained image FMs. Our code is provided at \url{https://github.com/ubc-tea/DNE-foundation-model-fairness}.


\end{abstract}
{%
  \let\thefootnote\relax%
  \footnotetext{Corresponding author: Xiaoxiao Li}%
  \let\thefootnote\svthefootnote%
}
\section{Introduction}
\label{intro}
Using pre-trained models or encoders to convert complex input data into vector representations is widely used in computer vision and natural language processing (NLP)~\cite{zhou2023comprehensive,jaiswal2024emergence}. This process transforms the original data into a representative low-dimensional hidden space, preserving information for downstream tasks. With the advancement of Foundation Models (FM), services like Google MedLM~\cite{singhal2023towards}, \href{https://www.voyageai.com/}{Voyage.ai}, and \href{https://openai.com/blog/chatgpt}{ChatGPT} provide data embedding services, eliminating the need for specialized hardware or extensive training. Their effectiveness is especially notable in medical fields, addressing data scarcity and hardware constraints. Clinics can enhance these models with minimal effort by fine-tuning custom classifiers on compact embedding output from FM application programming interface (APIs), which provides machine learning (ML) service as a function without users to train their own models from scratch.  However, the inherent biases in the APIs' training data and their model potentially harm marginalized groups by perpetuating gender, race, and other biases, a problem that has been exposed in the NLP field~\cite{schramowski2022large, you2024calibrating}. The biased embedding can compromise models' performance on minority groups~\cite{glocker2021algorithmic,jin2024fairmedfm}. Therefore, it is essential to develop new solutions to attain fairness in medical image embedding from pre-trained FM API.

Studies have been done to address the fairness issues in classification problems using ML models. These methods can be categorized into three approaches: 1)\underline{Model-based} strategies update or remove bias-related model parameters to mitigate bias, for example, using adversarial methods~\cite{adeli2021representation,wadsworth2018achieving, lim2023biasadv}; or prune parameters that are significant for sensitive attributes (SAs)~\cite{wu2022fairprune}; or update model to minimize the mutual information between target and SA representations using disentanglement learning~\cite{deng2023fairness,chen2023fedsoup}. However, the pretrained FM API services offer users very limited control over the model parameters~\cite{jones2024causal,larrazabal2020gender}. Thus model-based approaches are infeasible under the constraints.
2)\underline{Prediction calibration-based} methods analyze the prediction probability distribution of a classifier and apply different thresholds to each subgroup to reduce the discrepancy among subgroups~\cite{hardt2016equality,menon2018cost, sagawa2019distributionally}. These methods need custom thresholds for various tasks, classes, and groups, which limits their generalizability. Moreover, a significant amount of validation data is needed to determine the thresholds, rendering them less efficient in medical imaging applications where data scarcity is an unavoidable problem.
3)\underline{Data-based} strategies alleviate bias at the pre-training stage. 
Re-distribution and re-weighting methods~\cite{puyol2021fairness,sagawa2019distributionally} address unfairness by adjusting the balance of subgroups. The effectiveness of redistributed training data is limited because it can only affect the classification head, not the pre-trained FM API embedding encoder.
Recent data editing methods show strong ability to reduce disparity among subgroups by removing sensitive information from the input images~\cite{lim2023biasadv,wachinger2021detect}. However, can leave the model unchanged, \textit{they fail to address the underlying bias encoded within the black-box model, e.g., FM API}. Moreover, these approaches either depend on disease labels and require significant computational costs. Yao \emph{et al.} introduce an image editing method independent on targeted downstream tasks via sketching~\cite{yao2022improving}, but this method suffers from subpar performance as not learnable.  

In this work, we take two unique properties of pretrained FM API into consideration. Firstly, we aim to emphasize the need to maintain the flexibility of using FM without limiting it to a specific task while maintaining good utility, thus focusing on the learnable \underline{data editing-based} method that is independent of the downstream task. We choose to edit on the image space rather than on the FM API embeddings since the former provides better control on medical image fidelity and enables interpretability on the applied editing (as shown in App.~\ref{app:visualization}). 
Secondly, given the constraint against modifying parameters in the pre-trained API, we commit to not changing the FM API model parameters, even with black-box access to the API. To the best of our knowledge, no effective and universal~\footnote{We refer `universal' as debiasing API embeddings for various classification tasks.} method has been available to address this important, timely, but challenging research question: \textit{How to remove bias on medical images when using their embeddings from a pre-trained FM API for various classification tasks?}

To answer this question, we propose the \underline{D}ebiased \underline{N}oise \underline{E}diting, \imgedit, where the resulted edit can be shared across all subjects to eliminate SA-related information for various disease classification tasks. Specifically, we start by using a pre-trained SA classifier, trained with the same FM API embeddings.
Then, we optimize a set of learnable parameters (referred \imgedit{} noise) to be added to the images by confusing the SA classifier.
Furthermore, we introduce a greedy zeroth-order optimization strategy, \gezo, when APIs restrict gradient propagation in the black-box setting. Lastly, we apply this \imgedit{} noise to input images to generate fair embeddings and achieve unbiased disease classification across various disease tasks. Extensive evaluations on disease classification tasks show our method's effectiveness in promoting fairness while preserving utility.
\section{Method}
\subsection{Problem Setting}
\label{method:setting}
\begin{figure}[t]
    \centering
    \includegraphics[width=0.95\linewidth]{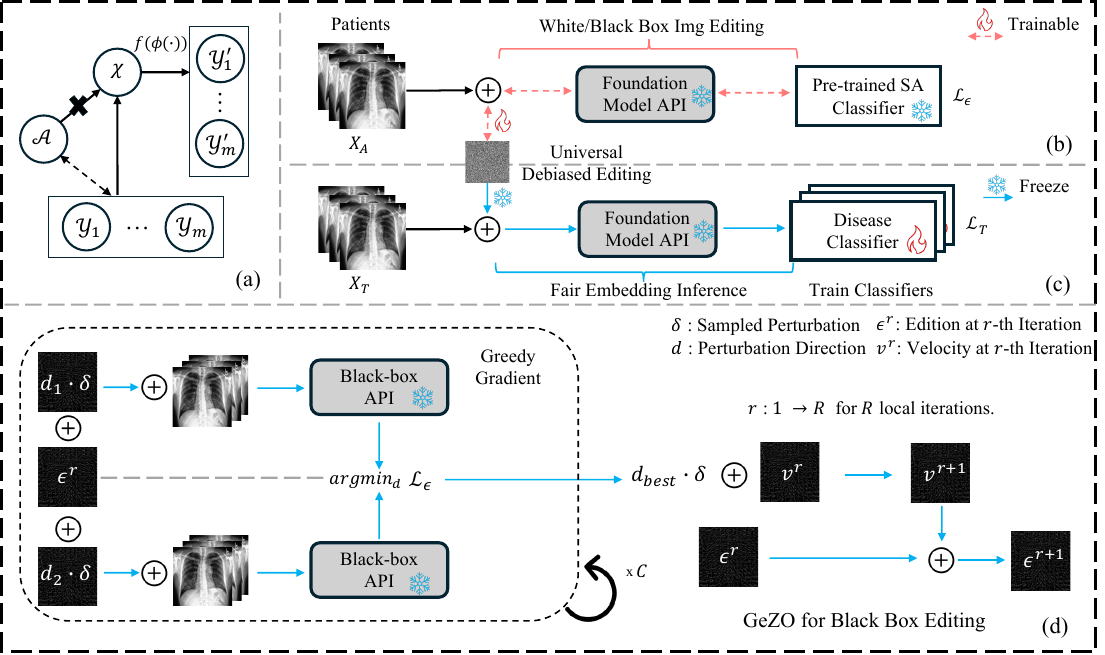}
    \caption{Overview of debias noise editing pipeline. (a) We eliminate the spurious correlation by breaking the connection between SA $\mathcal{A}$ and image $\mathcal{X}$, ensuring the model relies solely on disease-related information $\mathcal{Y}$. (b) Training \imgedit{} noise deceives the pre-trained SA classifier in $\mathcal{A}$ using a frozen FM API and SA classifier, suitable for both white-box and black-box API scenarios based on gradient accessibility. (c) We demonstrate the use of \imgedit{} noise: users augment their images with this noise, extract embeddings via the FM API, and proceed to train fair disease classifiers. (d) The (G)reedy (Z)eroth-(O)rder (GeZO) black-box editing method, selects the optimal perturbation via the \textit{Greedy Gradient} process when gradients is inaccessible. It tracks the best perturbation using velocity in each local iteration to update the \imgedit{} noise $\epsilon$.}
    \label{fig:main}
\end{figure}

This section outlines the problem of fairness in a binary medical image classification task. Firstly, we define key variables: input images $x \in \mathcal{X}$, binary disease labels $y\in \mathcal{Y} = \{0,1\}$, and sensitive attribute $a \in \mathcal{A} = \{1,...,|\mathcal{A}|\}$ (\ie~$|\mathcal{A}|=2$ for gender with male and female). As patients may not have sensitive and disease labels simultaneously, we denote images with sensitive attribute labels as $ \{x_i,a_i\}^N = [X_A,A]$ and images with disease label as $\{x_j,y_j\}^M = [X_T,Y]$, where $N$ and $M$ are the number of samples. Then we denote the FM API as \(\phi: \mathcal{X} \rightarrow \mathcal{Z}\), to obtain the image embedding \(z = \phi(x)\); the disease classifier as \(f: \mathcal{Z} \rightarrow \mathcal{Y}\) and SA classifier as  \(g: \mathcal{Z} \rightarrow \mathcal{A}\). \textit{We summarize all notations in App.~\ref{app:notation}.}

\noindent\textbf{Fairness Issue:} Fig.~\ref{fig:main} (a) shows there exists an association between sensitive attributes $\mathcal{A}$ and targets $\mathcal{Y}$ (dashed line). In the Empirical Risk Minimization (ERM)~\cite{vapnik1991principles}, the model $f(\phi(\cdot))$ superficially consider $\mathcal{A}$ as a proxy of $\mathcal{Y}$, causing the issue of fairness. This phenomenon persists even when there is an equal number of data points for each class in both \(\mathcal{Y}\) and \(\mathcal{A}\), as demonstrated in Sec.~\ref{exp:setting}. Our approach aims to address these disparities using \textit{debiased noise edit}.
\subsection{Debiased Noise Editing on Image for Fair Disease Classification}\label{method:img_edit}
In this section, we present our image editing approach aimed at reducing bias via learning a noise tensor with the same dimensionality as the image, which is called \imgedit{} noise. Then, we explain how to use this \imgedit{} noise for fair disease classification.

\noindent \textbf{Debiased Noise Editing.} Given the spurious correlation between SA $\mathcal{A}$ and disease $\mathcal{Y}$, we aim to remove this link through intervention on $\mathcal{X}$ by adding minimal editing vector $\epsilon$, known as \imgedit{}, concealing the spurious correlations while preserving image utility. As shown in Fig.~\ref{fig:main} (b), we first obtain a pre-trained SA classifier for medical images by either leveraging an existing one (if available) or using the collected $[X_A,A]$ to train the classifier $g$ with cross-entropy loss. Then, we freeze the classifier and leverage gradient ascent to learn the minimal \imgedit{} noise $\epsilon$, which is constrained by certain threshold to preserve image fidelity. $\epsilon$ is trained to  deceive the SA classifier $g$ without compromising useful information by solving the following optimization problem:
\begin{align}
\label{eq:adv}
    \mathcal{L}_{\epsilon} = -\left[\frac{1}{N} \sum_{i=1}^N \mathcal{L}_\texttt{CE}(a_i, g(\phi(x_i+\epsilon)))\right] + \lambda||\epsilon||_2
\end{align}
where $\lambda$ regularizes the magnitude of $\epsilon$. For example, larger $\lambda$ enforces the small L2-norm of the $\epsilon$ in the process of gradient descent. Sec.~\ref{exp:abl} explores its effect in detail. Importantly, we also offer a zero-order strategy for black-box APIs, as detailed in Sec.~\ref{method:zero}.

\noindent \textbf{Fair Disease Classification} Having obtained the \imgedit{} noise $\epsilon$ to conceal the SA $\mathcal{A}$, we then  freeze and add it to disease classification images $X_T$. As shown in Fig.~\ref{fig:main} (c), this action guarantees that the FM generates a fair embedding devoid of sensitive information, denoted as $\hat{z}_i = \phi(x_i+\epsilon)$. Finally, these fair embeddings are leveraged by the subsequent disease classifier $f$ for prediction. To train the classifier $f$, we employ a cross-entropy loss:
\begin{equation}
\label{eq:disease}
    \mathcal{L}_{T} =  \frac{1}{M} \sum_{i=1}^M \mathcal{L}_\texttt{CE}(y_i, f(\phi(x_i+\epsilon)),
\end{equation}
where classifier $f$ is the only trainable parameter and $M$ is the size of $X_T$.
\subsection{Greedy Zero-order Optimization for Black-box FM API}
\label{method:zero}

Zero-order (ZO) optimizations~\cite{chen2017zoo,liu2020primer}, which avoid the need for gradient computations, offer distinct advantages in optimizing black-box models. However, these methods often exhibit slower convergence rates, particularly in training large FM models. In contrast, our proposed \imgedit, concentrating on the input space, offers a mitigation of this training efficiency issue of ZO optimization. 

Inspired by ZO-SGD~\cite{spall1992multivariate} and recent MeZO~\cite{malladi2024fine} that employ in-place perturbation updates, we propose a G(reedy) Z(ero-)O(rder) (\gezo) optimization specifically for efficient \imgedit. \gezo{} employs in-place perturbations~\cite{spall1992multivariate,malladi2024fine} and greedily updates with the gradient sign that achieves global optimal loss, thereby accelerating the optimization process. As shown in Fig.~\ref{fig:main} (d), the core procedure of \gezo{} is the \textit{Greedy Gradient} in the right box, where the gradient of \imgedit's objective (Eq. ~\ref{eq:adv}) is estimated for each local iteration $r \in \{1, ..., |R|\}$ in \gezo. \textit{Greedy Gradient} takes the \imgedit{} noise from the current local iteration ($\epsilon^r$) and estimates its gradient by continuously adding minor perturbations ($\delta$) in different directions (\emph{e.g.}, $d_1$ and $d_2$) to it, \emph{i.e.}, $d_1 \cdot \delta$ and $d_2 \cdot \delta$. It then greedily selects the best direction perturbation ($d_\texttt{best} \cdot \delta$) that results in the smallest objective in this process. Like MeZO~\cite{malladi2024fine}, we integrate stochastic sampling for $C$ times. Each time we calculate the objective using a subset of training data to avoid sucking to the local optimum. Once the best direction and perturbation are selected for each local iteration $r$, we add $d_\texttt{best} \cdot \delta$ to the current velocity, $v^r$. This velocity keeps track of the accumulated updating direction and magnitude for $\epsilon$ for each local iteration $r$, where the initial value for velocity $v^1$ is 0. As shown in the right part of Fig.~\ref{fig:main} (d), $v^{r+1}$ is updated by adding the best perturbation returned by the \textit{Greedy Gradient}. Finally, the \imgedit{} noise $\epsilon^{r+1}$ is updated through adding the $v^{r+1}$. The local iterations $|R|$ is a hyper-parameter, where a smaller iteration increases the efficiency at the cost of the more biased gradient estimation. We investigate the effect of different $|R|$ in Sec.~\ref{exp:abl}. In actual implementation, we also introduced a momentum to update the velocity to accelerate the convergence. While the key idea is presented in this section,  we provide a detailed step-by-step algorithm box for \gezo{} in App.~\ref{app:black-box}.
\section{Experiement}

\subsection{Settings}
\label{exp:setting}
\noindent \textbf{Dataset.} To demonstrate the generalizability of \imgedit{}. we adopt the CheXpert dataset \cite{irvin2019chexpert}, a chest X-ray dataset with multiple disease labels, to predict the binary label for \textit{Pleural Effusion}, \textit{Pneumonia} and \textit{Edema} individually in chest radiographs. All the ablation studies are performed on \textit{Pleural Effusion} classification task for space limit. We take gender bias (male and female) as an example due to its broad impact on society and medical imaging analysis. To demonstrate the effectiveness of bias mitigation methods, we follow~\cite{deng2023fairness} to amplify the training data bias for each disease by (1) firstly dividing the data into different groups according to the SA; (2) secondly calculating the positive rate of each subgroup; (3) sampling subsets from the original training dataset and increase each subgroup's bias gap (more positive sample in a subgroup). Then, we sample testing data with the same procedure, but achieve an equal subgroup bias gap. The detailed data distribution is shown in Table~\ref{tab:combined-distribution} in App.~\ref{app:data}. 

\noindent\textbf{Evaluation metrics.} We use the classification accuracy to evaluate the utility of classifiers on the test set. To measure fairness,  we employ \textit{equal opportunity (EO)}~\cite{hardt2016equality} and \textit{disparate impact (DI)}~\cite{lipton2018does} metrics. EO aims to ensure equitable prediction probabilities across different groups, defined by the sensitive attribute $a^j$, for a given class $y^j$. It quantifies the disparity in true positive rates between groups: $\text{EO}_{Y=y_1} = P(\hat{Y} = y^1 | Y = y^1, A = a^1) - P(\hat{Y} = y^1 | Y = y^1, A = a^2)$,
where a smaller gap signifies greater equality of opportunity. DI evaluates the presence of indirect discrimination by measuring the ratio of positive predictions across different groups: $\text{DI} = \frac{P(\hat{Y} = 1 | A = a^1)}{P(\hat{Y} = 1 | A = a^2)}$.
DI closing to 1 indicates the minimal disparity in positive prediction rates between the groups. To align with EO metric, we employ the $|1-\textsc{DI}|$ to quantify the fairness in percentage, with smaller values denoting greater fairness.

\subsection{Implementation Details}  
\noindent \textbf{Architecture.} In our implementation, all methods use the same architecture. To simulate the FM API, we utilize a publicly available self-supervised pre-trained ViT-base model~\footnote{\url{https://github.com/lambert-x/Medical_MAE}}, optimized on X-ray images, given its superior performance metrics among all other architectures~\cite{xiao2023delving}. Within our study, we fix the encoder of the ViT model and only fine-tune the classification head, simulating the configuration detailed in Sec.~\ref{method:setting}.

\noindent \textbf{Debiased Noise Editing.} For \imgedit, we first fine-tune the SA classifier, $g$, using the feature encoded by the FM ($\phi(x_i)$) with the training set of \textit{Pleural Effusion} in Table~\ref{tab:combined-distribution}. We update it using Adam with a learning rate (lr) of $0.0001$ for $50$ epochs. Second, we update the \imgedit{} through implementing Eq.~\ref{eq:adv} by initializing $\epsilon$ as a trainable PyTorch parameter with all entries initially set to zero. The parameter matches the input image dimensions of $224 \times 224$. The \imgedit{} can be optimized using classic gradient descent or \gezo. Finally, we fine-tune the disease classifier, $f$, with $\epsilon$ added on the input data as delineated in Eq.~\ref{eq:disease}. We update it using AdamW with lr of $1.25 \times 10^{-4}$ for $50$ epochs. We provide the visualization of the magnitude of \imgedit{} as an interpretation of \imgedit{} in App.~\ref{app:visualization} Fig.~\ref{fig:noise-map}, where \imgedit{} is smoothed by the Gaussian kernel. As shown, larger noises are added to the bottom to discriminate the gender-related features.

\noindent \textbf{GeZO.} Here, we keep all parts for \imgedit{} the same as above, except using \gezo{} to update the edit rather than the Adam, given that the gradient is not accessible. The only key parameter that we are interested in is the local iteration $T$, where we investigated it in Sec.~\ref{exp:abl}. Furthermore, the detailed implementation of the algorithm and hyperparameter setting for \gezo{} can be found in App.~\ref{app:black-box}.

\subsection{Comparison with Baselines}
\begin{table*}[t]
\centering
\caption{Comparision of binary prediction of \textit{Pleural Effusion, Pneumonia, and Edema}. We label the best performance in \textbf{bold} and the second-best performance with \underline{underline}. All the values are in percentage.}
\resizebox{\linewidth}{!}{%
\begin{tabular}{l|cccc|cccc|cccc}
\hline\hline 
\multirow{2}{*}{Diseases} & \multicolumn{4}{c}{Pleural Effusion} & \multicolumn{4}{c}{Pneumonia} & \multicolumn{4}{c}{Edema} \\
\cline{2-13}
 & $EO_n \downarrow$ &$EO_p \downarrow$& |1-DI| $\downarrow$ & Acc $\uparrow$ &  $EO_n \downarrow$ &$EO_p \downarrow$ & |1-DI| $\downarrow$ & Acc $\uparrow$ &  $EO_n \downarrow$ &$EO_p \downarrow$ & |1-DI| $\downarrow$ & Acc $\uparrow$ \\
\hline

ERM & 40.5 & 57.0 & 58.5 & \underline{72.9} & 70.0  & 70.0 & 74.5 & 59.6 & 42.0  & 39.0 & 42.0 & 74.5 \\
\hline
Sketch~\cite{yao2022improving} & 44.0  & 52.0 & 57.5 & 66.3 & 64.0  & 61.0 & 72.6 & 56.3 & 44.0  & \textbf{15.0} & \textbf{15.5} & 67.0 \\
\hline
Group DRO~\cite{sagawa2019distributionally} & 41.0  & 58.0 & 59.5 & 72.3 & 60.0  & 56.0 & 64.4 & 61.0 & 40.5  & 40.5 & 43.3 & 74.8 \\
\hline
Batch Samp.~\cite{puyol2021fairness} & 41.5  & 50.5 & 51.8 & 74.3 & 64.0  & 72.0 & 76.6 & 60.5 & 39.0  & 43.0 & 44.8 & \textbf{76.0} \\
\hline
BiasAdv~\cite{lim2023biasadv} & 39.0 & 54.5 & 58.2 & 71.1 & \underline{36.0} & 54.0 & 77.1 & 61.0 & 39.0 & 33.5 & 36.6 & 74.9 \\
\hline
\imgedit & \textbf{28.5}  & \textbf{23.0} & \textbf{25.6} & \textbf{75.1} & 38.0  & \textbf{25.0} & \underline{34.7} & \underline{61.3} & \textbf{27.5}  & 17.0 & 19.7 & \underline{75.6} \\

\imgedit-GeZO & \underline{37.0}  & \underline{25.0} & \underline{28.5} & 72.8 & \textbf{35.0}  & \underline{27.0} & \textbf{33.7} & \textbf{61.5} & \underline{34.0} & \underline{15.5} & \underline{17.2} & 75.1 \\
\hline
\hline
\end{tabular}
}
\label{tab:disease-baseline}
\end{table*}

\subsubsection{Quantitative Analysis.} 
Table~\ref{tab:disease-baseline} summarizes the performances among different diseases and baselines. The baselines include \textit{model-based} strategy, like biasAdv~\footnote{In BiasAdv's implementation, we treat FM API as a white box as it requires access to FM's gradient and there is no zeroth-order optimization for it yet.}~\cite{lim2023biasadv}; data-based strategies, like batch sampling~\cite{puyol2021fairness} with data re-distribution and sketch~\cite{yao2022improving} with data generation; and prediction calibration-based strategy, like Group DRO~\cite{sagawa2019distributionally}. The details of these baselines are introduced in Sec.~\ref{intro}. The results indicate that the \imgedit{} effectively balances fairness and utility. Taking \textit{Pleural Effusion} as the example to analyze, the white-box optimized \imgedit{} outperforms all baselines. The $EO_p$ and DI decrease over 25\% compared to all the baselines, an indication of less disparity between the two genders. Additionally, the utility not only maintains but also surpasses all other groups, e.g., \imgedit's accuracy (Acc) is higher than 2.2\% compared to ERM. Given we are using a balanced testing set, this increase is also a sign of better generalization. Similar for \textit{Pneumonia} and \textit{Edema}, where \imgedit{} and \imgedit-\gezo{} takes the best two performance for almost all entries. Although learned through black-box optimization, our \imgedit-\gezo{} achieves comparable performance to the standard optimization used in \imgedit. This demonstrates not only the validity of \gezo{} but also the efficiency afforded by the small number of edition parameters~\cite{malladi2024fine}. In \textit{Pneumonia}, \imgedit-\gezo{} even slightly surpasses the performance of \imgedit{} across most metrics. These findings affirm the efficacy of \imgedit, where it not only facilitates fair medical image embedding and training but also introduces better generalizability in downstream classification tasks. Furthermore, one trained \imgedit{} can be potentially applied to all other diseases of Chest Xray, where it consistently outperforms the baselines across three diseases in both fairness and utility.

\subsection{Ablation Studies}
\label{exp:abl}
\begin{figure}
    \centering
    \begin{minipage}{0.5\textwidth}
        \centering
        \includegraphics[width=\textwidth]{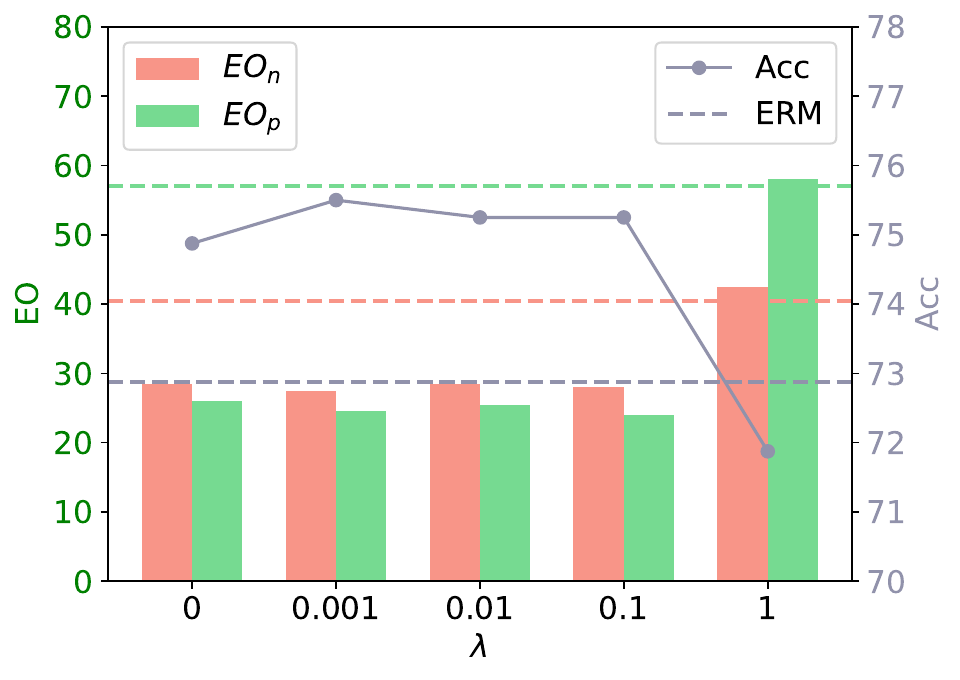}\\
        (a)
    \end{minipage}\hfill
    \begin{minipage}{0.5\textwidth}
        \centering
        \includegraphics[width=\linewidth]{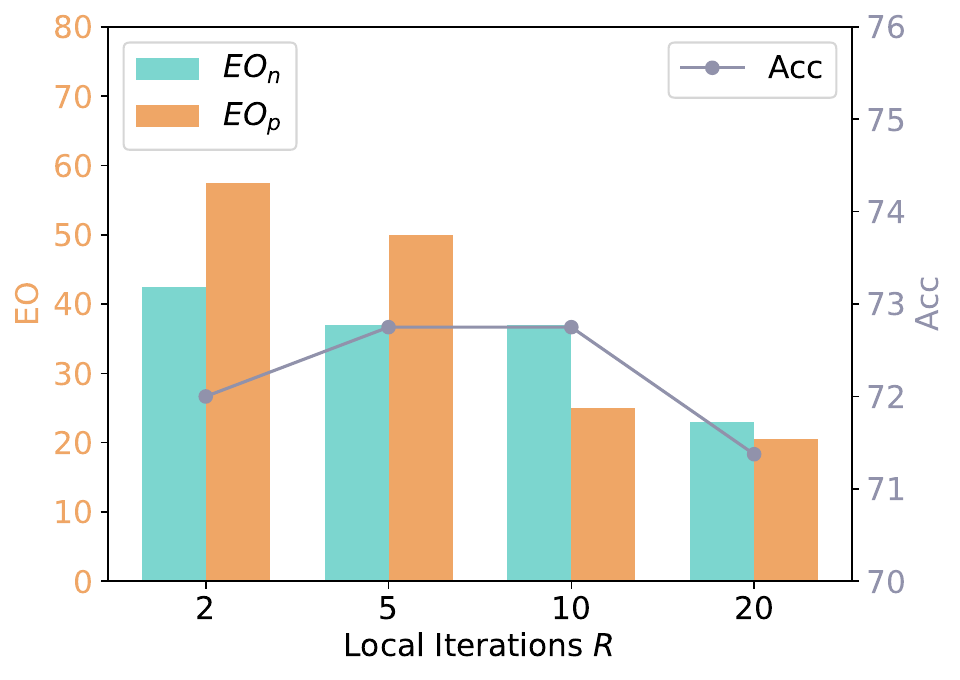} \\
        (b)
    \end{minipage}
    \caption{Ablation study of our methods: (a) Effect of different $\lambda$; (b) Effect of different local epochs using \gezo.}
    \label{fig:abl}
\end{figure}

\noindent \textbf{Effect of Regularization Coefficient.}
To investigate the effect of different regularization coefficients' ($\lambda$) effect, we vary $\lambda$ from 0 to 1. Fig.~\ref{fig:abl} (a) depicts the EO and the Acc for different $\lambda$s. The metrics for ERM are labeled as horizontal dashed lines for convenience.
As shown, both the fairness and utility metrics remain relatively stable for $\lambda < 1$, consistently surpassing the ERM baseline. However, as $\lambda$ increases to 1, we observe a marked decline in accuracy below the ERM benchmark, along with a notable increase in the EO metric for both classes.

\noindent \textbf{Effect of Local Iterations in \gezo.}
In black-box API, total local iterations $R$ in \gezo{} affect the optimization performance, wherein larger $R$ leads to more accurate optimization at the cost of efficiency. Here, we examine the effect of changing $R$, as shown in Fig.~\ref{fig:abl} (b). For fairness metrics, increasing $R$ from 2 to 20 significantly reduces the EO score for both classes, demonstrating a considerable debiasing impact with more local epochs. This effect is attributed to increased perturbation sampling that expands the search space with more local epochs, as introduced in Sec.~\ref{method:zero}. Meanwhile, accuracy remains relatively stable, with minor fluctuations between 72\% and 73\%. 
\section{Conclusion}
In this study, we address a crucial, yet under-explored aspect of health equity—the inherent bias in FM API's usage for classification—through the introduction of \textit{debiased noise editing}. \imgedit{} effectively masks bias-inducing pixels, enhancing fairness in API-generated embeddings. Furthermore, \gezo{} tackles the challenge of the inaccessibility of the gradient in black-box APIs by estimating gradients via perturbation. Future research will extend \imgedit{}'s application across various FM APIs and settings, aiming to solidify its role in promoting fairer machine learning practices.

\begin{credits}
\subsubsection{\ackname} This work is supported in part by the Natural Sciences and Engineering Research Council of Canada (NSERC), Compute Canada, and Vector Institute.

\subsubsection{\discintname}
The authors have no competing interests to declare that are
relevant to the content of this article.
\end{credits}

%
%

%
%
%
%
\bibliography{reference}

\begin{thebibliography}{29}
\providecommand{\natexlab}[1]{#1}
\providecommand{\url}[1]{\texttt{#1}}
\providecommand{\urlprefix}{}

\bibitem[{Adeli et~al.(2021)Adeli, Zhao, Pfefferbaum, Sullivan, Fei-Fei, Niebles, and Pohl}]{adeli2021representation}
Adeli, E., Zhao, Q., Pfefferbaum, A., Sullivan, E.V., Fei-Fei, L., Niebles, J.C., Pohl, K.M.: Representation learning with statistical independence to mitigate bias.
\newblock In: Proceedings of the IEEE/CVF Winter Conference on Applications of Computer Vision. pp. 2513--2523 (2021)

\bibitem[{Chen et~al.(2023)Chen, Jiang, Dou, Wang, and Li}]{chen2023fedsoup}
Chen, M., Jiang, M., Dou, Q., Wang, Z., Li, X.: Fedsoup: improving generalization and personalization in federated learning via selective model interpolation.
\newblock In: International Conference on Medical Image Computing and Computer-Assisted Intervention. pp. 318--328. Springer (2023)

\bibitem[{Chen et~al.(2017)Chen, Zhang, Sharma, Yi, and Hsieh}]{chen2017zoo}
Chen, P.Y., Zhang, H., Sharma, Y., Yi, J., Hsieh, C.J.: Zoo: Zeroth order optimization based black-box attacks to deep neural networks without training substitute models.
\newblock In: Proceedings of the 10th ACM workshop on artificial intelligence and security. pp. 15--26 (2017)

\bibitem[{Deng et~al.(2023)Deng, Zhong, Dou, and Li}]{deng2023fairness}
Deng, W., Zhong, Y., Dou, Q., Li, X.: On fairness of medical image classification with multiple sensitive attributes via learning orthogonal representations.
\newblock In: International Conference on Information Processing in Medical Imaging. pp. 158--169. Springer (2023)

\bibitem[{Glocker et~al.(2021)Glocker, Jones, Bernhardt, and Winzeck}]{glocker2021algorithmic}
Glocker, B., Jones, C., Bernhardt, M., Winzeck, S.: Algorithmic encoding of protected characteristics in image-based models for disease detection.
\newblock arXiv preprint arXiv:2110.14755  (2021)

\bibitem[{Hardt et~al.(2016)Hardt, Price, and Srebro}]{hardt2016equality}
Hardt, M., Price, E., Srebro, N.: Equality of opportunity in supervised learning.
\newblock Advances in neural information processing systems 29 (2016)

\bibitem[{Irvin et~al.(2019)Irvin, Rajpurkar, Ko, Yu, Ciurea-Ilcus, Chute, Marklund, Haghgoo, Ball, Shpanskaya et~al.}]{irvin2019chexpert}
Irvin, J., Rajpurkar, P., Ko, M., Yu, Y., Ciurea-Ilcus, S., Chute, C., Marklund, H., Haghgoo, B., Ball, R., Shpanskaya, K., et~al.: Chexpert: A large chest radiograph dataset with uncertainty labels and expert comparison.
\newblock In: Proceedings of the AAAI conference on artificial intelligence. vol.~33, pp. 590--597 (2019)

\bibitem[{Jaiswal et~al.(2024)Jaiswal, Liu, Chen, Wang et~al.}]{jaiswal2024emergence}
Jaiswal, A., Liu, S., Chen, T., Wang, Z., et~al.: The emergence of essential sparsity in large pre-trained models: The weights that matter.
\newblock Advances in Neural Information Processing Systems 36 (2024)

\bibitem[{Jin et~al.(2024)Jin, Xu, Zhong, Yao, Dou, Zhou, and Li}]{jin2024fairmedfm}
Jin, R., Xu, Z., Zhong, Y., Yao, Q., Dou, Q., Zhou, S.K., Li, X.: Fairmedfm: Fairness benchmarking for medical imaging foundation models.
\newblock arXiv preprint arXiv:2407.00983  (2024)

\bibitem[{Jones et~al.(2024)Jones, Castro, De~Sousa~Ribeiro, Oktay, McCradden, and Glocker}]{jones2024causal}
Jones, C., Castro, D.C., De~Sousa~Ribeiro, F., Oktay, O., McCradden, M., Glocker, B.: A causal perspective on dataset bias in machine learning for medical imaging.
\newblock Nature Machine Intelligence pp. 1--9 (2024)

\bibitem[{Larrazabal et~al.(2020)Larrazabal, Nieto, Peterson, Milone, and Ferrante}]{larrazabal2020gender}
Larrazabal, A.J., Nieto, N., Peterson, V., Milone, D.H., Ferrante, E.: Gender imbalance in medical imaging datasets produces biased classifiers for computer-aided diagnosis.
\newblock Proceedings of the National Academy of Sciences 117(23), 12592--12594 (2020)

\bibitem[{Lim et~al.(2023)Lim, Kim, Kim, Ahn, Shin, Yang, and Han}]{lim2023biasadv}
Lim, J., Kim, Y., Kim, B., Ahn, C., Shin, J., Yang, E., Han, S.: Biasadv: Bias-adversarial augmentation for model debiasing.
\newblock In: Proceedings of the IEEE/CVF Conference on Computer Vision and Pattern Recognition. pp. 3832--3841 (2023)

\bibitem[{Lipton et~al.(2018)Lipton, McAuley, and Chouldechova}]{lipton2018does}
Lipton, Z., McAuley, J., Chouldechova, A.: Does mitigating ml's impact disparity require treatment disparity?
\newblock Advances in neural information processing systems 31 (2018)

\bibitem[{Liu et~al.(2020)Liu, Chen, Kailkhura, Zhang, Hero~III, and Varshney}]{liu2020primer}
Liu, S., Chen, P.Y., Kailkhura, B., Zhang, G., Hero~III, A.O., Varshney, P.K.: A primer on zeroth-order optimization in signal processing and machine learning: Principals, recent advances, and applications.
\newblock IEEE Signal Processing Magazine 37(5), 43--54 (2020)

\bibitem[{Malladi et~al.(2024)Malladi, Gao, Nichani, Damian, Lee, Chen, and Arora}]{malladi2024fine}
Malladi, S., Gao, T., Nichani, E., Damian, A., Lee, J.D., Chen, D., Arora, S.: Fine-tuning language models with just forward passes.
\newblock Advances in Neural Information Processing Systems 36 (2024)

\bibitem[{Menon and Williamson(2018)}]{menon2018cost}
Menon, A.K., Williamson, R.C.: The cost of fairness in binary classification.
\newblock In: Conference on Fairness, Accountability and Transparency. pp. 107--118. PMLR (2018)

\bibitem[{Puyol-Ant{\'o}n et~al.(2021)Puyol-Ant{\'o}n, Ruijsink, Piechnik, Neubauer, Petersen, Razavi, and King}]{puyol2021fairness}
Puyol-Ant{\'o}n, E., Ruijsink, B., Piechnik, S.K., Neubauer, S., Petersen, S.E., Razavi, R., King, A.P.: Fairness in cardiac mr image analysis: an investigation of bias due to data imbalance in deep learning based segmentation.
\newblock In: Medical Image Computing and Computer Assisted Intervention--MICCAI 2021: 24th International Conference, Strasbourg, France, September 27--October 1, 2021, Proceedings, Part III 24. pp. 413--423. Springer (2021)

\bibitem[{Sagawa et~al.(2019)Sagawa, Koh, Hashimoto, and Liang}]{sagawa2019distributionally}
Sagawa, S., Koh, P.W., Hashimoto, T.B., Liang, P.: Distributionally robust neural networks for group shifts: On the importance of regularization for worst-case generalization.
\newblock arXiv preprint arXiv:1911.08731  (2019)

\bibitem[{Schramowski et~al.(2022)Schramowski, Turan, Andersen, Rothkopf, and Kersting}]{schramowski2022large}
Schramowski, P., Turan, C., Andersen, N., Rothkopf, C.A., Kersting, K.: Large pre-trained language models contain human-like biases of what is right and wrong to do.
\newblock Nature Machine Intelligence 4(3), 258--268 (2022)

\bibitem[{Singhal et~al.(2023)Singhal, Tu, Gottweis, Sayres, Wulczyn, Hou, Clark, Pfohl, Cole-Lewis, Neal et~al.}]{singhal2023towards}
Singhal, K., Tu, T., Gottweis, J., Sayres, R., Wulczyn, E., Hou, L., Clark, K., Pfohl, S., Cole-Lewis, H., Neal, D., et~al.: Towards expert-level medical question answering with large language models.
\newblock arXiv preprint arXiv:2305.09617  (2023)

\bibitem[{Spall(1992)}]{spall1992multivariate}
Spall, J.C.: Multivariate stochastic approximation using a simultaneous perturbation gradient approximation.
\newblock IEEE transactions on automatic control 37(3), 332--341 (1992)

\bibitem[{Vapnik(1991)}]{vapnik1991principles}
Vapnik, V.: Principles of risk minimization for learning theory.
\newblock Advances in neural information processing systems 4 (1991)

\bibitem[{Wachinger et~al.(2021)Wachinger, Rieckmann, P{\"o}lsterl, Initiative et~al.}]{wachinger2021detect}
Wachinger, C., Rieckmann, A., P{\"o}lsterl, S., Initiative, A.D.N., et~al.: Detect and correct bias in multi-site neuroimaging datasets.
\newblock Medical Image Analysis 67, 101879 (2021)

\bibitem[{Wadsworth et~al.(2018)Wadsworth, Vera, and Piech}]{wadsworth2018achieving}
Wadsworth, C., Vera, F., Piech, C.: Achieving fairness through adversarial learning: an application to recidivism prediction.
\newblock arXiv preprint arXiv:1807.00199  (2018)

\bibitem[{Wu et~al.(2022)Wu, Zeng, Xu, Shi, and Hu}]{wu2022fairprune}
Wu, Y., Zeng, D., Xu, X., Shi, Y., Hu, J.: Fairprune: Achieving fairness through pruning for dermatological disease diagnosis.
\newblock In: International Conference on Medical Image Computing and Computer-Assisted Intervention. pp. 743--753. Springer (2022)

\bibitem[{Xiao et~al.(2023)Xiao, Bai, Yuille, and Zhou}]{xiao2023delving}
Xiao, J., Bai, Y., Yuille, A., Zhou, Z.: Delving into masked autoencoders for multi-label thorax disease classification.
\newblock In: Proceedings of the IEEE/CVF Winter Conference on Applications of Computer Vision. pp. 3588--3600 (2023)

\bibitem[{Yao et~al.(2022)Yao, Cui, Li, and Gu}]{yao2022improving}
Yao, R., Cui, Z., Li, X., Gu, L.: Improving fairness in image classification via sketching.
\newblock arXiv preprint arXiv:2211.00168  (2022)

\bibitem[{You et~al.(2024)You, Min, Dai, Sekhon, Staib, and Duncan}]{you2024calibrating}
You, C., Min, Y., Dai, W., Sekhon, J.S., Staib, L., Duncan, J.S.: Calibrating multi-modal representations: A pursuit of group robustness without annotations.
\newblock arXiv preprint arXiv:2403.07241  (2024)

\bibitem[{Zhou et~al.(2023)Zhou, Li, Li, Yu, Liu, Wang, Zhang, Ji, Yan, He et~al.}]{zhou2023comprehensive}
Zhou, C., Li, Q., Li, C., Yu, J., Liu, Y., Wang, G., Zhang, K., Ji, C., Yan, Q., He, L., et~al.: A comprehensive survey on pretrained foundation models: A history from bert to chatgpt.
\newblock arXiv preprint arXiv:2302.09419  (2023)

\end{thebibliography}
\clearpage

\appendix
\section{Notation Table}
\label{app:notation}
\begin{table}[H]
\centering
\resizebox{\columnwidth}{!}{
\begin{tabular}{cl}
\toprule
Notations & Description \\ 
\hline \\[-1.8ex]
$x$, $\mathcal{X}$ & input image, input space\\

$y$, $\mathcal{Y}$ & disease label, label space\\

$a$, $\mathcal{A}$ & sensitive attribute label, sensitive label space \\

$z$, $\mathcal{Z}$ & embedding, embedding Space \\

$\phi: \mathcal{X} \to \mathcal{Z}$& freezed FM encoder \\

$f: \mathcal{Z} \to \mathcal{Y}$ & disease classifier \\

$g: \mathcal{Z} \to \mathcal{A}$ & sensitive attribute classifier \\

$\epsilon$ & universal edition \\

$\lambda$ & regularization coefficient \\

$T$ & disease target \\
\bottomrule
\end{tabular}
}

\caption{Notation Table}
\label{tab:notation}
\end{table}

\section{Visualization}
\label{app:visualization}

\begin{figure}
    \centering
    \includegraphics[width=0.9\linewidth]{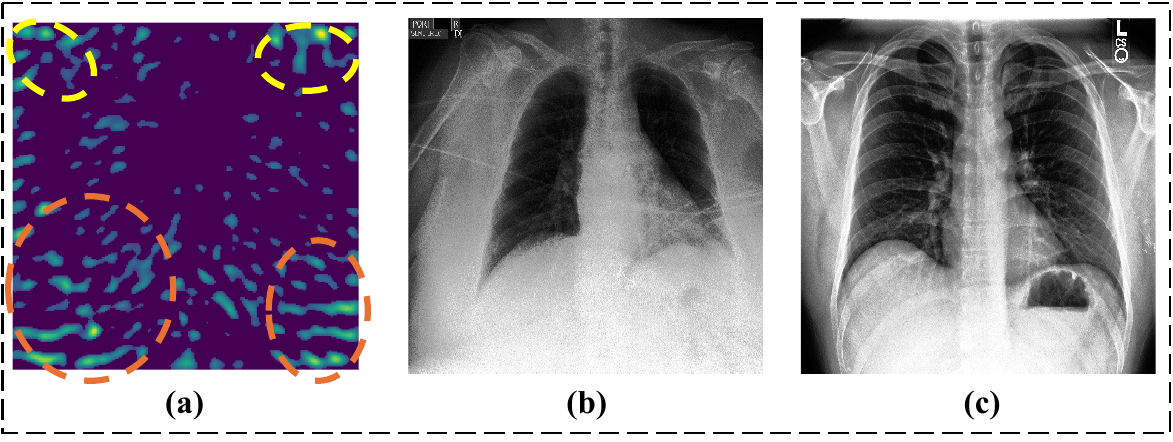}
    \caption{Visualization of chest X-rays and \imgedit{} noise patterns (with Gaussian smoothing applied) to interpret gender-discriminative image regions. (a) The normalized UDE noise map, with larger noise highlighted by brighter color, reveals gender-discriminative features. The large noise circled in orange corresponds to the breast. The large noise circled in yellow reflects artifacts on X-ray, such as text notations. (b) A female chest X-ray. (c) A male chest X-ray. }
    \label{fig:noise-map}
\end{figure}

\section{Data Distribution}
\label{app:data}
\begin{table}[h]
\centering
\caption{Training and Testing Set Distribution for \textit{Pneumonia, Edema, and Pleural Effusion}.}
\label{tab:combined-distribution}
\begin{tabular}{l|cc|cc|cc|cc|cc|cc}
\hline
\hline
   \multirow{3}{*}{Diseases}      & \multicolumn{4}{c|}{Pneumonia}              & \multicolumn{4}{c|}{Edema}                    & \multicolumn{4}{c}{Pleural Effusion}          \\
         \cline{2-13}
         & \multicolumn{2}{c|}{Negative} & \multicolumn{2}{c|}{Positive} & \multicolumn{2}{c|}{Negative} & \multicolumn{2}{c|}{Positive} & \multicolumn{2}{c|}{Negative} & \multicolumn{2}{c}{Positive} \\
         \cline{2-13}
         & M & F                 & M & F                & M & F                  & M & F                & M & F                 & M & F              \\
\hline
$\#$ Train Sample & 1500 & 150                     & 150  & 1500                   & 5000 & 500                     & 500  & 5000                   & 5000 & 500                     & 500  & 5000                   \\
$\#$ Test Sample  & 100  & 100                     & 100  & 100                    & 200  & 200                     & 200  & 200                    & 200  & 200                     & 200  & 200                    \\
\hline
\hline
\end{tabular}
\end{table}

\section{Greedy Zero-order Optimization}
\label{app:black-box}
\begin{algorithm}[H]
\caption{Greedy Zero-order (\gezo) Optimization}
\textbf{Setup: }{Local Iterations $R$, step size $s$, step size decay $k$, Edit from last global epoch $\epsilon_\texttt{epoch}$, sample times $C$, best loss $\mathcal{L}_\texttt{best}$, best direction $d_\texttt{best}$, momentum: $\mu$, batch size $B$}.

\begin{algorithmic}[1]
\Procedure{UpdateEditionPerEpoch}{$\epsilon_\texttt{epoch}$}:
\State $\epsilon^1 \gets \epsilon_\texttt{epoch}$, $s^1 \gets 0.01$,  $v^1 \gets 0$, $\mu \gets 0.9$, $\mathcal{L}_\texttt{best} \gets \infty$, $k \gets 0.95$

\For{$r = 1 \to R$}
    \State $d_\texttt{best},\mathcal{L}_\texttt{best} \gets$\Call{\text{GreedyGradient}}{$B,\epsilon^r,s^r$} \Comment{find the $d_\texttt{best}$ achieve $\mathcal{L}_\texttt{best}$}
    \If{$d_\texttt{best} \neq \text{None}$}
        \State $v^{r+1} \gets \mu \cdot v^{r} + d_\texttt{best}$
        \State $\epsilon^{r+1} \gets \epsilon^r + v^{r+1}$, $s^{r+1} \gets s^r$
        \Comment{Update $\epsilon$ with momentum}
    \Else
        \State $s^{r+1} \gets k \cdot s^r$
        \Comment{Reduce step size if no improvement}
    \EndIf
\EndFor
\State \Return $\epsilon^R$ \Comment{Return the updated $\epsilon_\texttt{epoch + 1}$}
\EndProcedure
\Procedure{GreedyGradient}{$B,\epsilon^r,s^r$}
 \State $d_\texttt{best} \gets \text{None}$
\For{$c = 1 \to C$}
    \Comment{Sample $C$ times}
        \State $\delta \gets \text{RandomPerturbation}() \cdot s_r$
        \Comment{Generate scaled perturbation}
        \For{$d \in \{-1, 1\}$}
            \State $\epsilon' \gets \epsilon^r + d \cdot \delta$
            \Comment{Apply perturbation in both directions}
                \State $\mathcal{L}_{\epsilon} = -\left[\frac{1}{B} \sum_{i=1}^B \mathcal{L}_\texttt{CE}(a_i, h(\phi(x_i+\epsilon')))\right] + \lambda||\epsilon'||_2$
                \Comment{One batch}
            \If{$\mathcal{L}_\epsilon < \mathcal{L}_\texttt{best}$}
                \State $\mathcal{L}_\texttt{best} \gets \mathcal{L}_\epsilon$, $d_\texttt{best} \gets d \cdot \delta$
                \Comment{Update best loss and direction}
            \EndIf
        \EndFor
    \EndFor
\State \Return $d_\texttt{best},\mathcal{L}_\texttt{best}$.
\EndProcedure
\end{algorithmic}
\label{alg:black-box}
\end{algorithm}

\end{document}